# Enhancing Bangla Fake News Detection Using Bidirectional Gated Recurrent Units and Deep Learning Techniques

Utsha Roy
Department of Computer Science
and Engineering
Daffodil International University
Dhaka, Bangladesh
utsha15-3094@diu.edu.bd

Mst. Sazia Tahosin
Department of Computer Science
and Engineering
Daffodil International University
Dhaka, Bangladesh
sazia15-3666@diu.edu.bd

Md. Mahedi Hassan
Department of Computer Science
and Engineering
Daffodil International University
Dhaka, Bangladesh
mhassan.cse@diu.edu.bd

Taminul Islam[†]
School of Computing
Southern Illinois University
Carbondale
IL, USA
taminul.islam@siu.edu

Fahim Imtiaz
Department of Computer Science
and Engineering
Daffodil International University
Dhaka, Bangladesh
fahim15-14447@diu.edu.bd

Md Rezwane Sadik
Department of Economics &
Decision Sciences
University of South Dakota
Vermillion, SD, United States
mdrezwane.sadik@coyotes.usd.edu

Yassine MALEH
LaSTI Laboratory
Sultan Moulay Slimane
University
Beni Mellal, Morocco
yassine.maleh@ieee.org

Rejwan Bin Sulaiman
Lecturer at Department of
Computer Science
Northumbria University
United Kingdom
rejwan.sulaiman@northumbria.ac.uk

Md. Simul Hasan Talukder
Bangladesh Atomic Energy
Regulatory Authority
Dhaka, Bangladesh
simulhasantalukder@gmail.com

## ABSTRACT

The rise of fake news has made the need for effective detection methods, including in languages other than English, increasingly important. The study aims to address the challenges of Bangla which is considered a less important language. To this end, a complete dataset containing about 50,000 news items is proposed. Several deep learning models have been tested on this dataset, including the bidirectional gated recurrent unit (GRU), the long short-term memory (LSTM), the 1D convolutional neural network (CNN), and hybrid architectures. For this research, we assessed the efficacy of the model utilizing a range of useful measures, including recall, precision, F1 score, and accuracy. This was done by employing a big application. We carry out comprehensive trials to show the effectiveness of these models in identifying bogus news in Bangla, with the Bidirectional GRU model having a stunning accuracy of 99.16%. Our analysis highlights the importance of dataset balance and the need for continual improvement efforts to a substantial degree. This study makes a major contribution to the creation of Bangla fake news detecting systems with limited resources, thereby setting the stage for future improvements in the detection process.

## 1 Introduction

In today's digital media age, the news media has an essential role in shaping public opinion and influencing decisions. But the rapid spread of information through online forums poses a problem: the spread of fake news [1]. The presentation of false claims or misleading information as truth can have far-reaching consequences such as distortion of facts, widespread misinformation, and even public unrest [2]. The detection of false information and has been resisted has emerged as a major challenge in information dissemination [3].

While considerable efforts have been made to address fake news, most studies have focused on major languages, leaving minority languages and communities with limited resources and solutions. One of those languages is Bangla, a language famous among millions [4]. However, the complexity of Bangla creates unique conditions for fake news detection. Linguistic diversity, idiomatic expressions, and cultural nuances provide a unique terrain with which traditional approaches can clash [5]. In response to this gap, our study embarked on a journey to investigate fake news in Bangladesh through deep learning methods. Our focus is on harnessing the power of a Gated Recurrent Unit (GRU), which is a form of neural network combined with other deep learning techniques [6]. This approach has shown remarkable success in a variety of natural language processing purposes, including translating languages to sentiment assessment.

A selection of articles specifically produced for the veracity and authenticity of Bangladeshi news forms the basis of our study. This dataset serves as a strong foundation for research and analysis and is a crucial tool for comprehending the intricacy of false news in Bangladesh. The dataset contains a variety of features, which allow us to capture news breakdowns variety of online distributions. We take a step-by-step approach to make our research accessible and discoverable. We start by processing the raw database and using WordClouds to visualize the most frequently used words. Finally, we engage in data preparation that includes processing such as term removal and correction. This ensures that our model gets clean and relevant information without unwanted noise.

The creation and implementation of deep learning models are the primary focus of our research. In the process of our study, we made use of many different models, namely 1D Convolutional Neural Networks, Long Short Term Memory Networks, and a variety of hybrid models for deep learning. Through intensive research and analysis, we found a paradigm that showed the highest accuracy and reliability in false case detection in the case of Bangladesh. Furthermore, we examine the importance of addressing class inequality, which is a common challenge in fake news detection. We use regression methods to correct for this inequality and assess the impact on model performance. Not only are we doing research, but we are also making efforts to determine the most efficient methods by which to cut down on false negatives and false positives, which are two essential markers of the dependability of false detection systems.

As a result of fusing deep learning and annotation databases, our study entails a thorough attempt to enhance Bangladeshi fake news. We want to contribute to a larger conversation about preventing misinformation online during this time by delving into the specifics of Bangladeshi fake news identification and maximizing the two-track GRU model's capability. By bridging the gap between cutting-edge technology and Bangladesh's particular issues in false news identification, we illustrate our journey, discoveries, and insights in the pages that follow. Here is the contribution of our work:

1. Study the effectiveness of the bidirectional Gated Recurrent unit (GRU) in determining the accuracy of Bangla news articles.
2. Improving the accuracy of the model in identifying authentic and fake information, addressing the problem of class inequality using alternative modeling techniques.
3. Bridging the gap between advanced intensive learning and the unique challenges of the Bangla language, providing practical solutions to important challenges.

## 2   Literature Review

We provide a brief overview of key projects, embark on a journey through existing research, and present a summary of their findings.

Zobaer Hossain et al. [7] introduced a dataset of 50,000 Bangla news articles for automated fake news detection. They explored traditional and neural network methods, highlighting SVM's success with a 91% F1-score, while neural models like CNN and BiLSTM performed less effectively. The dataset is focused on Bangla news articles. Risul Rasel et al. [8], by merging freshly acquired fake news data with earlier datasets, we produced a Bangla false news dataset. The study made use of a range of deep neural network (LSTM, Bi-LSTM, CNN, LSTM-CNN, Bi-LSTM-CNN), Transformer (Bangla-BERT, m-BERT), machine learning (LR, SVM, KNN, MNB, Adaboost, DT), and deep learning (LSTM, Bi-LSTM, CNN) models. Notably, CNN came out as the highest performer, earning accuracy rates of 95.9% respectively while improving recall for false news data. The dataset comprised 4678 distinct news items, and feature extraction employed techniques like Count Vectorizer and TF-IDF Vectorizer. The models exhibited improved accuracy (1.4% to 3.4%) when tested against existing datasets. Md Hossain et al. [9] presented strategies to enhance Bangla fake news detection by addressing imbalanced datasets. They applied these methods to BanFakeNews, a skewed dataset with 97% majority instances, using oversampling techniques like Random Oversampling, SMOTE, and ADASYN. The authors achieved a 93.1% F1-score with SMOTE and a 79.1% F1-score with Stacked Generalization. Notably, without these techniques, baseline models would yield a 67.6% F1-score. Although the paper focuses solely on Bangla, it marks a significant step toward advancing fake news detection. Md Ali et al. [10] proposed a more effective machine learning method for identifying bogus news in Bengali. Their focus was on achieving higher recognition rates, utilizing a dataset merging authentic and fake news categories. Employing diverse feature extraction methods and machine learning algorithms, they attained an impressive 98.08% accuracy through their Python-based implementation. However, the study points out that the feasibility of detecting Bengali fake news from social media warrants further exploration. Liza Fahmida et al. [11] demonstrate a multilingual fake news detection algorithm capable of concurrently detecting bogus information in both Bengali and English. They use TF-IDF and N-Gram Analysis to extract features and compare the output of six machine learning techniques. With 93.29% accuracy and a 0.93 F1 score, the model performs well when utilizing the Linear Support Vector Classification approach. Although the dataset includes articles in both Bengali and English, the study's key contribution is a unique approach that can be used to news portals in both languages. However, the unavailability of the dataset is a limitation. Sourav Saha et al. [12] introduce a transfer learning-based approach employing BERT's masked LM technique for Bengali fake news detection. Leveraging pre-trained BERT, they attain a accuracy and precision rate of 95%, surpassing models like LSTM, SVM, NB, and CNN. The study's methodology involves employing SVM and MNB, utilizing ReLU activation, and implementing masked LM (MLM) preprocessing. Notably, the paper's scope is limited to detecting fake news in Bengali. Manal Iftikhar and Arshad Ali et al. [13] present a machine learning model for automated fake news detection. They gather information from many internet sources and use SVM, Naive Bayes, and LSTM classification approaches to distinguish between authentic and false news. Their method produces outstanding results, with a decision tree classifier on the ISOT false news dataset obtaining 99% accuracy. Additionally, ensemble techniques like Support Vector Machine (SVM) and Random Forest (RF) are used. In particular, the suggested SVM plus RF model outperforms the current accuracy by 2.3%. Investigating Covid-19-related English fake news, the authors employ ensemble techniques, combining Roberta and BERT, achieving a remarkable 98% accuracy. Leela et al. [14] provide an example of a false news detection system that uses the Logistic Regression (LR) method and compares its accuracy to the Support Vector Machine (SVM) technique. When identifying bogus news, LR performs better than SVM (91.68%), with an accuracy of 95.12%. a 311-sample dataset that will be used in the study. The detection process used both LR and SVM algorithms, with LR showing to be more accurate. However, the paper lacks details about the dataset source and the system's applicability to diverse datasets or domains. Tamim Al Mahmud and Antara Mondal [15] introduce an innovative technique for Bengali news classification via machine learning and deep learning models. Their approach involves collecting over 20K structured data, applying various models, and selecting the best one. The method uses many measures, including as precision, recall, accuracy, and F-measure, to assess the performance of the model. Notably, they achieve 87.12% accuracy using the support vector machine (bag of words) model. Although the paper's contribution lies in an accuracy-enhancing technique, a limitation is noted in their prior work's limited model scope. The models used include gated recurrent units, long short-term memory, random forests, linear support vector classifiers, and support vector machines. Elias Hossain et al. [14] uses a mix of deep learning and machine learning to investigate the detection of fake news in Bangla. Using a dataset of 57,000 Bangla news articles, they research algorithms including Bi-LSTM with Glove and FastText as well as Gated Recurrent Unit (GRU). The study highlights the superiority of Deep Learning over Traditional Machine Learning. The Bi-LSTM model achieves a robust 96% accuracy, while the GRU model scores 77%. The paper further includes a comparative analysis of existing work and experimental insights. The proposed system showcases adaptability for real-time news classification. However, the study's scope is solely Bangla fake news detection, limiting its application to other languages. Iftikhar Ahmad, Muhammad Yousaf, Suhail Yousaf, and Muhammad Ovais offer a machine learning ensemble technique for automatic classification of news articles as real or counterfeit [15]. The study explores distinct textual properties and employs ensemble methods and linguistic feature sets across various domains. Using datasets like ISOT Fake News Dataset and Kaggle's offerings, the authors evaluate linear SVM and ensemble techniques. The results showcase the ensemble learner approach's superiority over individual learners, with the bagging classifier (decision trees) topping at 94% accuracy. Notably, boosting classifier (XGBoost) excels in precision with 95.25%, while bagging classifier excels in recall at 0.942. XGBoost's efficient error identification principle drives its superior performance. The study's limitation lies in evaluating algorithms across diverse domains within a single dataset.

## 3 Proposed Methodology

Figure 1 depicts the processes in our research methods that can help Bangladesh better identify bogus news. We used a variety of data related to the different media. We collected data from a public dataset that was collected through Bangladeshi news articles. This data set was thoroughly reviewed, where we cleaned up the text and removed unnecessary terms to ensure the sample would have clear and relevant information. To understand the common words, we visualize them with WordClouds. We explore the world of deep learning, building and training models with pre-generated data. These models consist of a bidirectional gated recurrent unit (GRU) [16], 1D convolutional neural networks (CNN) [17], Long short-term memory (LSTM) [18], and their combinations. Through intensive analysis, we found the best models for false information found in Bangla language specifics. In addition, we use permutation methods to address the issue of class heterogeneity to ensure a balanced representation and improve overall model performance. Our approach is not just an introduction; it includes finding ways to reduce false positives and false negatives. This holistic approach, which is a combination of deep learning methods and linguistic expertise, enables us to effectively contribute to combating misinformation in the Bangladeshi linguistic context.

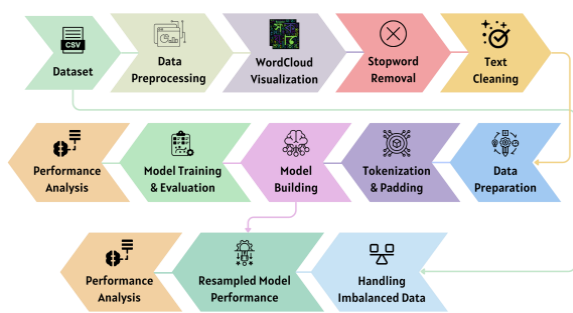

**Figure 1: Our Proposed Methodology**

### 3.1 Dataset Description

The dataset for this research was collected from the open-source Kaggle [19] database. Four primary files make up our suggested database, and each one helps us understand fake news in Bangla:

1. Authentic-48K.csv and Fake-1K.csv: These files contain essential columns, including the article ID, news publisher's domain, publication date, category, headline, news content, and label (1 for authentic, 0 for fake). These files form the core of our dataset, offering a diverse selection of authentic and fake news articles.
2. LabeledAuthentic-7K.csv and LabeledFake-1K.csv: These files provide additional context and insight into the news articles. They include columns such as source (the entity that can verify the news claim), relation (related or unrelated), and F-type (type of fake news, present only in LabeledFake-1K.csv). These files augment our dataset by introducing elements crucial for in-depth analysis.

Below is a summarized table detailing the columns found in each of these files:

**Table 1: Dataset Column Description**

| Column Name | Description of the Column |
|---|---|
| articleID | News ID |
| domain | Site name of news publisher |
| date | Publication Date |
| category | Category of the news |
| source | Source of the news |
| relation | Related or Unrelated |
| headline | Headline of the news |
| content | Article of the news |
| label | 1 as authentic, 0 as fake |
| F-type | Fake news type (only in LabeledFake-1K.csv) |

### 3.2 Data Preprocessing

The goal of our data pre-processing is to transform the unprocessed textual input into a structure suitable for analysis and model training. The first step is to extract the necessary "content" and "label" columns from real and fake news databases. Then, we proceed to combine the extracted datasets to create a combination of training and testing datasets. After setting up our database, we proceeded with typographical corrections, such as removing additional letters and punctuation. These activities are performed to maximize the appropriateness of the words in the text. The use of WordCloud visualization provides a visual representation that highlights most occurrences of words and facilitates understanding of language structure. Tokenization is the process of transforming words into mathematical vectors, allowing them to be combined without hard work in deep learning models [20]. Using this comprehensive data processing technique enhances our ability to perform in-depth analysis and enables our models to more accurately identify misinformation in Bangla.

### 3.3 WordCloud Visualization

The role of highly visualizing words in our training and testing datasets, as shown in Figures 2 and 3, is important in our analysis. This approach provides visual identification of important words in the information in the report, making it easier to identify themes and patterns present in the language used [21]. Valuable insights about implicit and unique features can be gained. The diagram is a valuable tool to guide our data creation process. The following allows for and also gives a realistic picture of explicit linguistic development in Bangla. Using this simple technique, we can make connections between raw data and critical analysis, thereby facilitating informed decision-making in the development of our false-positive analysis strategy.

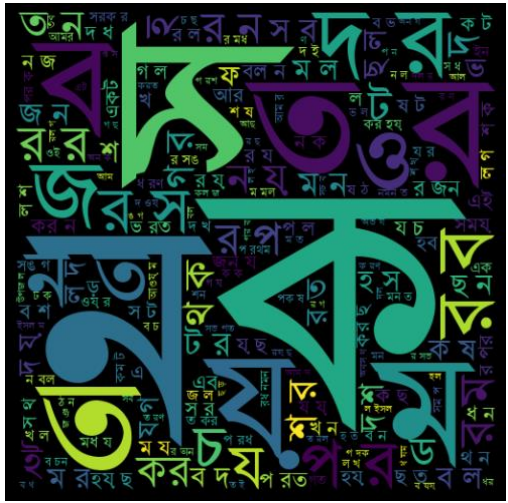

**Figure 2: Word Visualization in Train Dataset**

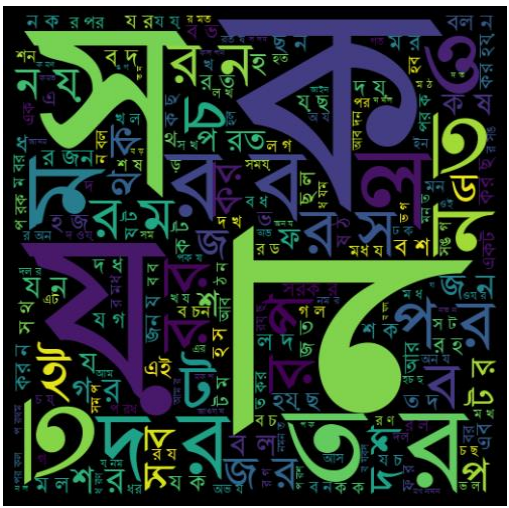

**Figure 3: Word Visualization in Test Dataset**

### 3.4 Stopword Removal

In our efforts to improve the quality of the text, we acknowledge the existence of terminology. The aforementioned terms, despite their widespread use, generally lack significant linguistic context and do not make sense in the classification system To overcome this issue, we use terms that stand use a special library called "Stopwords Bengali (BN)", which we got from GitHub [22]. The library provides an extensive collection of Bengali characters specially preserved to correspond with the grammar of our language. By integrating these libraries, we effectively eliminate redundant characters, thus improving the efficiency of our data processing. To enhance the logic of our text classification efforts, we eliminate common punctuation, as shown in the examples below:

<p align="center">ওরা, কখনও, ছিলেন, জন, তোমার, থাকবে etc.</p>

### 3.5 Text Cleaning

In the process of enhancing the dataset, we engage in an important process called text correction. This strategy includes two main strategies designed to enhance the quality and relevance of written content. Initially, news is put into a filtering system in which only those above a certain word count limit of 100 words are retained The use of filtering techniques assures that we are committed to information quality needs and informs. In addition, a text correction technique is applied to systematically change the content by removing extraneous characters such as line breaks, tabs, special marks, etc. Numeric characters and lines have been removed from common symbols and specific Bengali characters from the text. By performing a comprehensive storage process, we ensure that our dataset will be optimized for comprehensive analysis. This refinement enables our model to effectively assess Bangla news authenticity with high accuracy.

### 3.6 Data Preparation

For model training and testing during data preparation, we didn't split it randomly. We extract the essential "content" and "label" columns from the authentic and fake news databases, where there are a total of 49,000 articles, which we use as a training dataset. The same for the testing dataset, where we merged the label_authentic and label_fake csv files and got a total of 8,000 articles.

### 3.7 Tokenization and Padding

Tokenization and padding play an important role in our data usage. Tokenization is the process of converting textual information into numeric sequences and provides the linguistic processing capabilities of our model [23]. The relationship between written language and mathematical symbols is established by assigning a specific numerical value to each word. Padding is a technique for ensuring that all sequences maintain a constant length, thus making them more likely to converge when training patterns [24]. The process of converting expressions into numeric representations and checking that sequence length moves so constantly enables our models to probe the complex linguistic nuances of text. This stage is important to ensure that our models can successfully draw knowledge from the content and accurately distinguish between real and fake Bangla news.

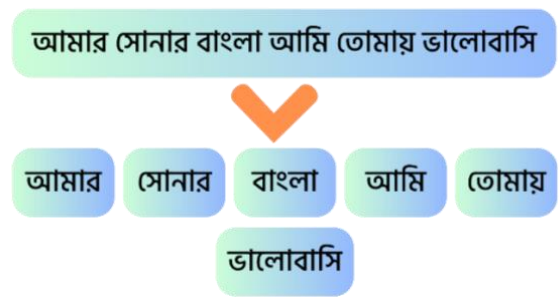

**Figure 4: Example of Tokenization**

### 3.8 Models

Our study uses several methods to comprehend the reliability of Bangla news. The 1D Convolutional Neural Network (CNN), Long Short-Term Memory (LSTM), Bidirectional Gated Recurrent Unit (GRU), and composite variations like CNN + LSTM, CNN + GRU, and CNN + LSTM + Global Max Poling 1D are all included in our analysis of deep learning models [25]. We use a variety of models to take advantage of their ability to capture language variations, patterns, and context. This comprehensive analysis helps us in our goal of creating a reliable Bangla fake news detection system.

*3.8.1 Bidirectional Gated Recurrent Unit.* We begin by using a bidirectional GRU model, which effectively captures succeeding patterns by combining data from both directions. An embedding layer, a bidirectional GRU layer, and a global Average Pooling layer are all included in the model. Dense layers of ReLU activation enable nonlinear changes and reduce dropout overfitting. The output layer consists of a sigmoid activation resulting in binary classification results.

*3.8.2 1D Convolutional Neural Network.* Our research extends to 1D Convolutional Neural Network (CNN) models. It includes an embedding layer, a 1D Convolutional layer, a MaxPooling layer, and smoothing for feature extraction. Later dense layers and dropout methods improve the generalizability of the model and prevent overfitting.

*3.8.3 Long Short-Term Memory.* This model follows suit, making the most of its sequential memory storage. The embedding layers are connected to LSTM layers, followed by Dense layers with dropouts for regularization. *The sigmoid* function in the output layer helps in binary classification.

*3.8.4 CNN + LSTM.* A hybrid CNN-LSTM approach combines the advantages of convolutional and iterative layers. This example combines the embedding, convolutional, and LSTM layers using the unique properties of both architectures. The outputs of the Convolutional and LSTM layers are combined before moving on to the Dense layer.

*3.8.5 CNN + GRU.* Another hybrid approach, this model combines Convolutional and Gated Recurrent Unit (GRU) levels. Convoluted layers remove features, while GRU layers require sequential dependencies. The particles of the two materials are correlated before entering the Dense layers.

*3.8.6 CNN + LSTM + Global Max Poling 1D.* We continue to innovate by modeling combining Convolutional, LSTM, and GlobalMaxPooling1D layers. This hybrid architecture combines the feature extraction capabilities of Convolutional layers, sequential analysis of LSTM, and global reference capture of GlobalMaxPooling1D.

## 3.8 Proposed Model

Our proposed architecture uses a bidirectional gated recursive unit (GRU) to predict the accuracy of Bangla news content. This architecture takes use of GRU's bidirectional nature, enabling the model to effectively capture context in both directions. Figure 5 shows the architecture graphically.

The model begins with an embedding layer that converts textual input into numerical representation. A form of recursive layer adopted for memory holding capacity is the bidirectional GRU layer at the heart of the architecture. This layer stores information about the preceding and following terms, improving the understanding of the context of the model. The GlobalAveragePoling1D layer stores context and subsequent Dense layers introduce nonlinearity. Dropout levels reduce overfitting and improve generalization. The final output layer uses sigmoid activation, which allows the classification of valid or deceptive information in binary classification. This example is the basis of our attempt to provide a robust fake detection method for the Bangla language domain.

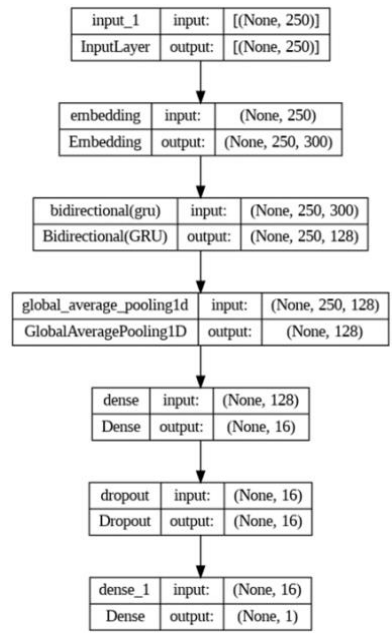

**Figure 5: Proposed Architecture of Bidirectional Gated Recurrent Unit**

## 4 Experimental Analysis

We evaluate the robustness of our model using important measures including accuracy, precision, recall, and F1 scores. These measurements serve as a bright compass that leads us through the convoluted routes of our model's performance. Exploring these metrics reveals vital information about the capabilities of our model.

**Accuracy.** Accuracy measures the proportion of cases that were correctly predicted and all recorded cases in the data set. Calculated as follows [26]:

$$Accuracy = \frac{TP + TN}{Total\ Instances} \quad (1)$$

**Precision.** Precision refers to the ability of the model to detect true positives among all predicted positives. The formula is given as follows [27]:

$$Precision = \frac{TP}{TP + FP} \quad (2)$$

**Recall.** Recall refers to the ability of the model to correctly identify all actual positives among predicted positives. The formula is explained as follows [28]:

$$Recall = \frac{TP}{TP + FN} \quad (3)$$

**F1 score.** The F1 score combines precision and recall, providing a balanced assessment of model performance. It can be calculated using this formula [29]:

$$F1\ score = 2 * \frac{Precision * Recall}{Precision + Recall} \quad (4)$$

These metrics provide a more thorough understanding of the model's ability to flip between true positives (TP), true negatives (TN), false positives (FP), and false negatives (FN), providing a more complete view of the model's efficacy.

### 4.1 Result Analysis

Here we conducted a comprehensive evaluation of various models using key performance metrics. The metrics encompass accuracy, precision, recall, and F1 score, providing a comprehensive view of the models' performance.

**Table 2: Before Balancing the Dataset**

| Models | Accuracy | Precision | Recall | F1 score |
|---|---|---|---|---|
| **Proposed Model** | **99.12%** | **99.28%** | **99.85%** | **99.56%** |
| 1D CNN | 99.00% | 99.07% | 99.91% | 99.49% |
| LSTM | 97.59% | 97.59% | 99.99% | 98.78% |
| CNN + LSTM | 98.84% | 98.91% | 99.91% | 99.41% |
| CNN + GRU | 98.61% | 99.04% | 99.53% | 99.29% |
| CNN + LSTM + GlobalMaxPooling1D | 98.71% | 99.02% | 99.66% | 99.34% |

We assess our model and provide the results in Table 2 before balancing the dataset. The amazing 99.12% accuracy, 99.28% precision, 99.85% recall, and 99.56% F1 score of the bidirectional GRU model is rather remarkable. The 1D CNN model exhibits adequate accuracy, with an accuracy of 99.00%, with equal precision, recall, and F1 score values of 99.07%, 99.91%, and 99.49%. The LSTM model has an accuracy of 97.59% with precision, recall, and F1 score values of 97.59%, 99.99%, and 98.78%, respectively. Other models, including CNN + LSTM, CNN + GRU, and CNN + LSTM + GlobalMaxPooling1D, do as well in these criteria. The combination of these findings demonstrates how well our suggested methodology captures bogus news from Bangladesh.

**Table 3: After Balancing the Dataset**

| Models | Accuracy | Precision | Recall | F1 score |
|---|---|---|---|---|
| **Proposed Model** | **99.16%** | **99.4%** | **99.74%** | **99.57%** |
| 1D CNN | 99.06% | 99.24% | 99.8% | 99.52% |
| LSTM | 97.49% | 98.02% | 99.43% | 98.72% |
| CNN + LSTM | 98.9% | 99.22% | 99.65% | 99.44% |
| CNN + GRU | 98.45% | 99.13% | 99.29% | 99.21% |
| CNN + LSTM + GlobalMaxPooling1D | 98.58% | 99.16% | 99.38% | 99.27% |

Table 3 displays the results of our model analyses using a balanced set of data. Metrics like accuracy, precision, recall, and F1 scores may be used to evaluate the model's performance in a balanced setting. The two-stage GRU model has a 99.16% accuracy, with accuracy values of 99.4%, 99.74%, and 99.57%. When recall and F1 score values are included, the accuracy of the 1D CNN model is 99.06%, and the corresponding accuracy, recall, and F1 score values are 99.24%, 99.8%, and 99.52%. The LSTM model's respective accuracy, precision, recall, and F1 score are 97.49%, 98.02%, 99.43%, and 98.72%. On these metrics, the CNN + LSTM, CNN + GRU, and CNN + LSTM + GlobalMaxPooling1D models all perform well. The findings show that our models perform better when used with a balanced dataset, which considerably improves their ability to detect false news in Bangladesh.

After analyzing the results, we balance the data sets to provide a comprehensive picture of the performance of our models and show the confusion matrix (CM) [30] and ROC-AUC curve [31] for each model figure 6 - 11 provides these diagrams, including true positive, true negative, false positive, and false negative classifications of models Obviously. The ROC-AUC curves illustrate the trade-off between true positive rate and false positive rate, and demonstrate the model's ability to distinguish between valid and false positives These graphs provide in-depth insight into how models move effective and flexible in dealing with the fake journalism in Bangla.

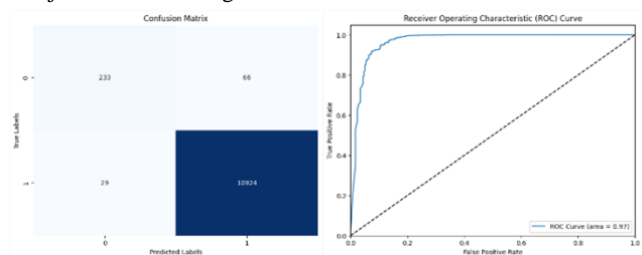

**Figure 6: CM and ROC-AUC Curve of Bidirectional GRU**

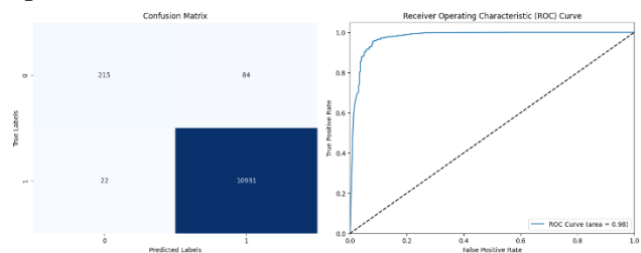

**Figure 7: CM and ROC-AUC Curve of 1D CNN**

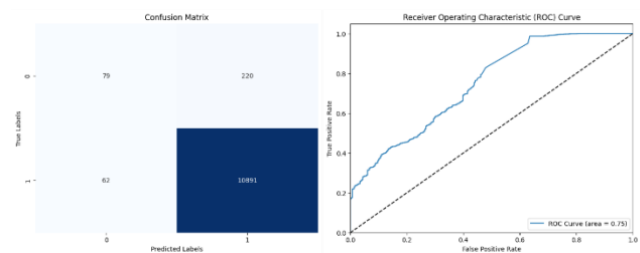

**Figure 8: CM and ROC-AUC Curve of LSTM**

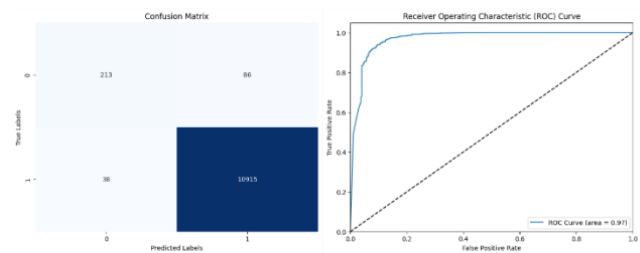

**Figure 9: CM and ROC-AUC Curve of CNN + LSTM**

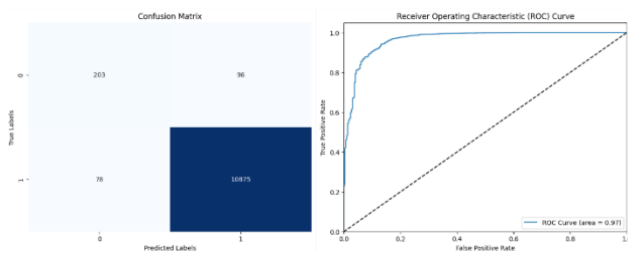

**Figure 10: CM and ROC-AUC Curve of CNN + GRU**

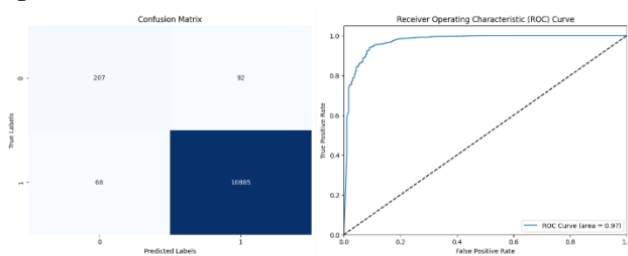

**Figure 11: CM and ROC-AUC Curve of CNN + LSTM + GlobalMaxPooling1D**

## 5   Discussion

In this study, we looked at how to handle fake news in Bangla. Our work tried to close the fake news detection gap by concentrating on the usage of resource-free language in light of the fast dissemination of false information. We attempted to improve the accuracy and reliability of automated Bengali false news recognition by introducing a carefully generated dataset and using multiple innovative deep learning models, including 1D CNN, LSTM, hybrid architecture, and bidirectional GRU. Bidirectional GRU gave outstanding results in accuracy, precision, recall, and F1 scores after a thorough performance evaluation, and confusion matrices and ROC-AUC curves before and after balancing the dataset provided further insights into the model's performance and discriminative power. This study compares some previous study that are given below in table 4.

**Table 4: Comparison with other studies**

| Ref | Contribution | Model | Dataset | Accuracy |
|---|---|---|---|---|
|  | The authors have utilized Word2Vec algorithm to train on a large Bangla textual dataset | LSTM | 48k authentic news, 1k false news, 7k genuine news with labels, and 1k fake news with labels | accuracy of 91% |
|  | This study addresses the lack of research on low-resource languages like Bangla | (CNN), (CNN-LSTM), (BiLSTM) | g 4678 distinct news data by combining newly collected fake news data | 95.9%, 95.5%, 95.3% |
|  | The proposed model utilizes machine learning and deep learning algorithms, to detect fake news | Bi-LSTM | 57,000 Bangla news articles collected from various news portals of Bangladesh. | 96% |
|  | The model has competitive performance over existing models and can be used as an automation tool to check the authenticity of Bangla news | deep learning | 48,678 authentic articles and 1,299 fake articles in Bangla. | 98.29% |
| | Our Study | Bidirectional GRU | collected from the Kaggle Authentic-48K, LabeledAuthentic-7K | 99.16% |

This research not only contributes to an expanding body of knowledge in the fight against misinformation but also provides the insights and perspectives needed to address similar challenges in other languages with limited resources.

## 6   Conclusion

Our research made a concerted effort to address the spread of fake news in the Bangla context. We hope to improve fraud detection using advanced deep learning algorithms by creating and analyzing a huge amount of data. The effectiveness of our models in identifying false positives is shown by the favorable outcomes we saw in accuracy, precision, recall, and F1 scores. However, this journey does not end here. In future work, we recognize the importance of balancing data and intend to expand our data set by collecting more pseudo-events assuring a more representative and balanced training environment. This enhancement will strengthen the flexibility of our models and enable practical real-world applications. As we continue to improve our methods, we hope to not only help reduce misinformation in Bangla but also help detect fake news in a wider range of languages.

## ACKNOWLEDGMENTS

Not applicable.